\documentclass[10pt,twocolumn,letterpaper]{article}

\newif\ifarxiv
\arxivtrue

\usepackage{iccv}
\usepackage{times}
\usepackage{epsfig}
\usepackage{graphicx}

\usepackage[numbers]{natbib}

\usepackage{amsmath}
\usepackage{amssymb}
\usepackage{algorithm}
\usepackage{algorithmic}

\usepackage{amsfonts}
\usepackage{amsbsy}
\usepackage{multirow}
\usepackage[titletoc,toc,title]{appendix}

\usepackage{booktabs}
\usepackage{graphicx}
\usepackage{subcaption}
\usepackage{amsthm}

\newcommand{\PP}{\mathbb{P}}
\newcommand{\QQ}{\mathbb{Q}}
\newcommand{\EE}{\mathbb{E}}
\newcommand{\XX}{\mathcal{X}}
\newcommand{\YY}{\mathcal{Y}}
\newcommand{\xx}{\mathbf{x}}
\newcommand{\yy}{\mathbf{y}}

\renewcommand{\emph}{\textit}

\usepackage{authblk}

\usepackage[breaklinks=true,bookmarks=false]{hyperref}

\iccvfinalcopy 

\def\iccvPaperID{****} 

\def\corresponding{Corresponding author (mikbinkowski \texttt{at} gmail \texttt{dot} com).}

\begin{document}
\title{Batch weight for domain adaptation with mass shift}

\author[1, 2]{Miko{\l}aj Bi{\'n}kowski\thanks{\corresponding{}}}
\author[1, 3]{R Devon Hjelm}
\author[1, 4]{Aaron Courville}

\affil[1]{Mila, Universit\'e de Montr\'eal}
\affil[2]{Imperial College London}
\affil[3]{Microsoft Research}
\affil[4]{CIFAR Fellow}

\maketitle




\begin{abstract}
Unsupervised domain transfer is the task of transferring or \emph{translating} samples from a source distribution to a different target distribution.
Current solutions unsupervised domain transfer often operate on data on which the modes of the distribution are well-matched, for instance have the same frequencies of classes between source and target distributions.
However, these models do not perform well when the modes are not well-matched, as would be the case when samples are drawn independently from two different, but related, domains.
This \emph{mode imbalance} is problematic as generative adversarial networks (GANs), a successful approach in this setting, are sensitive to mode frequency, which results in a mismatch of semantics between source samples and generated samples of the target distribution.
We propose a principled method of re-weighting training samples to correct for such mass shift between the transferred distributions, which we call \emph{batch-weight}. We also provide rigorous probabilistic setting for domain transfer and new simplified objective for training transfer networks, an alternative to complex, multi-component loss functions used in the current state-of-the art image-to-image translation models. The new objective stems from the discrimination of joint distributions and enforces cycle-consistency in an abstract, high-level, rather than pixel-wise, sense. Lastly, we experimentally show the effectiveness of the proposed methods in several image-to-image translation tasks.   
\end{abstract}

\section{Motivation}
Recent developments originating from Generative Adversarial Networks~\citep[GANs,][]{goodfellow} allow generation of high quality images, often hardly distinguishable from the real images~\citep{karras2017, biggan}. 
Adversarial methods have also been successfully applied in conditional image generation, where generated samples are obtained as functions of some prior data. When the latter is also composed of images, the problem is also called \emph{image-to-image (domain) transfer} or \emph{style transfer}. In this scope, adversarial objectives are often combined with other loss functions to ensure desirable properties of the transfer networks. These include variants of \emph{cycle consistency} loss, originally proposed in CycleGAN~\citep{cycle-gan} and developed further by \citet{munit, aug-cycle, drit}, leading to current state-of-the-art results in several image-to-image transfer problems.

Notwithstanding these notable results, not much attention has been paid towards understanding of unsupervised domain transfer from probabilistic point of view. Although original CycleGAN assumes deterministic transfer, later works identified the necessity of learning non-deterministic \emph{many-to-many} mappings to account for features that might be present only in one of the considered domains. A common assumption made in such setting, sometimes directly~\citep[e.g., ][]{munit}, but often implicitly, is the existence of latent variable that covers the \emph{shared semantics} between the domains of interest. Learning the transfer function can then be decomposed to learning deterministic encoders to and stochastic decoders from such latent space.

This view, however, does not take into account the distributional differences that may exist between the domains being matched. Since probability mass is preserved trough encoders and decoders, every mode in source domain covers the same share of the distribution as its representations in latent space and the target domain do in their respective distributions. However, we do not necessarily want to match modes that consist the same shares of data. 

For example, consider the task of transferring between handwritten digits~\citep[MNIST,][]{mnist}, and Street View House Numbers~\citep[SVHN,][]{svhn}. 
These datasets are independently sampled but share semantics expressed in the digit classes, styless (e.g., seven with or without a cross), etc which we wish to correctly transfer.
Since digits in MNIST are evenly distributed, we expect the \emph{correct} transfer function to produce samples in which zeros cover approximately 10\% of all generated ones. 
Yet, the distribution of the ``correct" transfer function (one that maintains the digit class) would be different from the actual SVHN distribution, where ones cover around 20\% of the data. 
Therefore such a transfer function would not be optimal in the Optimal Transport sense, i.e. it would not minimize any divergence between reference and generated distributions. 
Given such correct transfer-generator, a \emph{good} discriminator would need to be insensitive to the disparity of mode (i.e., digit) frequencies between the source and target distributions.
If not, it would provide a gradient signal to the transfer-generator that would encourage it to alter some modes to account for the missing mass of ones in its output.

We will call the described issue a \emph{mode-mass imbalance}.

This issue demonstrates the view that shared semantics can be modelled through a latent variable is, in general, invalid. However, the described problem is inherent in all GAN-based approaches to domain transfer, since GAN discriminators are always trained to estimate some kind of divergence e.g., \emph{Jensen-Shannon}~\cite{goodfellow}; \emph{Wasserstein distance}~\cite{wgan, gulrajani}; \emph{Maximum Mean Discrepancy}~\cite{li2017mmd, mmdgans} between reference distribution and the generated samples.
For these reasons, we propose \emph{batch weight} as a solution to the issue caused by mass-preserving property of optimal transport in the context of domain transfer. Batch weight aims to re-balance samples within each batch to account for differences in the reference and generated distributions.

\section{Related Work}
The need to correct for mode shift between distributions of interest has been widely studied in machine learning. 

\subsection{Supervised learning}
In supervised learning, it is often assumed that distributions $p(x)$ and $q(x)$ of the independent variable on the training and tests are different, but the conditionals $p(y|x)$ and $q(y|x)$ are equal. Such situation is known as \emph{covariate shift} or \emph{sample selection bias} and has been addressed in multiple works \citep{shimodaira2000, zadrozny2004, huang2006,gretton2008,sugiyama2008}. 

The complimentary setting when $p(x|y) = q(x|y)$, termed \emph{label shift}, has also been studied \citep{zhang2013}. In more recent work, \citet{lipton2018} consider label shift correction for black box predictors.

\subsection{Importance sampling}
The problem of changing probability measure in empirical setting has has long been studied in \emph{importance sampling} theory. This general technique has been used in estimating properties of distribution available indirectly through another distribution. 

Importance sampling is often applied in variance reduction problems. In such, one re-weights the available sample so that the variance of the estimated quantity under new distribution is lower than with respect to the original one.
\subsection{Domain transfer}

The distribution shift has undergone some limited study in the context of domain transfer. \citet{Cohen2018DistributionML} empirically showed that the use of distribution-matching loss functions in domain transfer leads to issues when modes in target domain are under- or over-represented as compared to the source. \citet{diesendruck2018importance} also identifies the problem of distribution mismatch in generative modelling, however the proposed re-balancing function is provided only in case when relation between source and target distribution is available (directly or indirectly).

State-of-the-art unsupervised image-to-image translation models \citep{munit,aug-cycle,drit} assume that we are given samples $X$ and $Y$ drawn from two domains $\XX$ and $\YY$ according to some distributions $\PP_x$ and $\QQ_y$, and seek (possibly random) generator functions $G_{xy}:\XX\to\YY$ and $G_{yx}:\YY\to\XX$ so that generated distributions $G_{xy}\#\PP_x$ and $G_{yx}\#\QQ_y$\footnote{$f\#\PP$ denotes \emph{push-forward} measure of $\PP$ through function $f$.} match with $\QQ_y$ and $\PP_x$ in \emph{consistent} way. Numerous techniques have been developed to ensure that the generated samples are consistent with their sources and that the transfer is invertible. Most of them are based on simple yet powerful \emph{cycle-consistency loss} \citep{cycle-gan},
\begin{align}
    L^{cyc}(x, y) &= \|G_{yx}(G_{xy}(x)) - x\|_1
    \nonumber\\
    & + \| G_{xy}(G_{yx}(y)) - y \|_1,
    \label{eq:cycle-consistency}
\end{align}
along with the GAN objective \citep{goodfellow}; \citet{munit,drit,bicyclegan} combine as many as five different loss components to train the generator functions.

Note that in the presence of mode-mass imbalance, the assumption that generators should minimize the GAN objective violates the consistency between sources and targets. It can be shown that this assumption together with cycle consistency imply that our samples are drawn from marginals of the same distribution $\PP_{xy} = \QQ_{xy}$ on $\XX\times\YY$, i.e. $\PP_x = \EE_y \PP_{xy}, \QQ_y = \EE_x\PP_{xy}$. 

More complicated loss functions used in domain transfer still seek the generators that mimic the conditionals $\PP_{y|x}$ and $\QQ_{x|y}$, given marginals $\PP_x$ and $\QQ_y$. Although they do not imply the equality of the joint distributions $\PP_{xy}, \QQ_{xy}$ being searched (which correspond to the correct transfer), they do imply equality of their marginals, $\PP_x = \QQ_x$ and $\PP_y = \QQ_y$, which is impossible in the presence of mode-mass imbalance.

Domain transfer models often assume existence of underlying \emph{shared semantics}~\citep[e.g.][]{munit} modelled as underlying latent variable $U$ that consists common features of $X$ and $Y$. In such scenario, one aims to train encoders $Enc(U|X), Enc(U|Y)$ and decoders $Dec(X|U), Dec(Y|U)$, and transfer between domains through $U$. This, however, does not account for the possible mode-mass imbalance between $X$ and $Y$: even non-deterministic encoders and decoders preserve probability mass between $X$ and $U$, and $U$ and $Y$. Therefore, the assumption that such $U$ exists is, in general, invalid. This inherent issue of domain transfer has been noted by \citet{lavoie2018unsupervised}.

\section{Formulation}
In this section we propose a framework to perform the unsupervised domain transfer task in the presence of mode-mass imbalance. We formulate it separately for one-sided and two-sided transfer that involve reweighting one or two domains, respectively.

\subsection{One-sided batch weight}
Assume that $\PP_x$ and $\QQ_y$ are \emph{source} and \emph{target} measures on domains $\XX$ and $\YY$, respectively. We assume that correct domain transfer from $\XX$ to $\YY$ can be represented by joint distribution $\PP_{xy}$ such that the marginal $\PP_y = \EE_x\PP_{xy}$ covers the target distribution 
\begin{equation}\label{eq:supports1}
    \textrm{supp} \QQ_y \subset \textrm{supp} \PP_y.
\end{equation}
This assumption is much weaker than equality $\QQ_y = \PP_y$ which most domain transfer models implicitly assume.

We would like to learn to transfer (possibly non-deterministically) by training a generator function $G:\XX\to\YY$ that mimics the conditional $\PP_{y|x}$. 
Let $D\in\mathcal{L} = \mathrm{Lip1}(\YY, \mathbb{R})$ be a Wasserstein discriminator (we will stick to the Wasserstein framework, however similar arguments can be derived in general for any divergence).

The Wasserstein GAN optimizes the following loss function,
\begin{equation} \label{eq:wasserstein}
    \inf_{G}\sup_{D\in\mathcal{L}}\EE_{X\sim\PP_x}[D(G(X))] - \EE_{Y\sim\QQ_y} [D(Y)],
\end{equation}
which is equivalent to,
\begin{equation}
    \inf_{G}\sup_{D\in\mathcal{L}} \EE_{Y\sim\PP^G_y}[D(Y)] - \EE_{Y\sim\QQ_y} [D(Y)]),
\end{equation}
where $\PP^G_y = G\#\PP_x$ is a push forward measure of $\PP_x$ through $G$. 

This optimization suffers from the problem of mode-mass imbalance, as in general we \emph{do not} want $\PP^G_y$ to match with $\QQ_y$. However we do expect $\PP^G_y$ to cover all the modes of $\QQ_y$, as $\PP_y$ does due to the assumption \ref{eq:supports1}. If this is true, then the Radon-Nikodym derivative $\frac{d\QQ_y}{d\PP^G_y}$ exists and, 
\begin{align} \label{eq:expectations}
    \EE_{Y\sim\QQ_y}[D(Y)] &= \EE_{Y'\sim\PP^G_y}\left[D(Y')\frac{d\QQ_y}{d\PP^G_y}(Y')\right] \nonumber\\
    &= 
    \EE_{X\sim\PP_x}\left[D(G(X))\frac{d\QQ_y}{d\PP^G_y}(G(X))\right].
\end{align}
$\frac{d\QQ_y}{d\PP^G_y}(G(\cdot))$ is an unknown function, and therefore the last expression in the above equation cannot be obtained directly. However, we may try to estimate it using e.g. neural network $W$,
\begin{equation} \label{eq:w}
    \inf_{W\in\mathcal{W}}\left(
    \EE_{X\sim\PP_x}[D(G(X)\cdot W(X)] - \EE_{Y\sim\QQ_y}[D(Y)]\right)^2,
\end{equation}
where $\mathcal{W} = \left\{W:\EE_{X\sim\PP_x}[W(X)] = 1, W \geq 0\right\}$. Such constraint can easily be enforced by a softmax layer computed over samples in the batch.

Problems \ref{eq:wasserstein} and \ref{eq:w} together motivate the following optimization criterion for Wasserstein batch-weighted domain transfer:
\begin{equation} \label{eq:bw}
    \inf_{G,W}\sup_{D}\left(\EE_{X\sim\PP_x}[D(G(X))\cdot W(X)] - \EE_{Y\sim\QQ_y} [D(Y)]\right)^2.
\end{equation}
Although optimization of the Wasserstein objective is technically equivalent to optimization of its square, in practice it is more convenient to use standard WGAN loss. Therefore the training procedure optimizes slightly different losses for weighting and generator networks.
The proposed procedure for batch-weighted domain transfer is shown in Algorithm \ref{alg:bw} in \ifarxiv Appendix \ref{a:one-sided}\else supplementary material\fi.

\subsubsection{Possible issues} \label{s:issues}
Samples from modes in the source domain that are underrepresented in the target domain might be transferred poorly if too low weights are assigned to them by the weighting network. This problem essentially stems from the fact that we are weighting the \emph{generated} samples, not the \emph{target} ones, which biases the generator so that it values generated samples according to their frequency in target domain $\QQ_y$, even though we care about quality of $\PP_y^G$, which stems from $\PP_x$.

\begin{figure*}
    \centering
    \includegraphics[width=.8\textwidth]{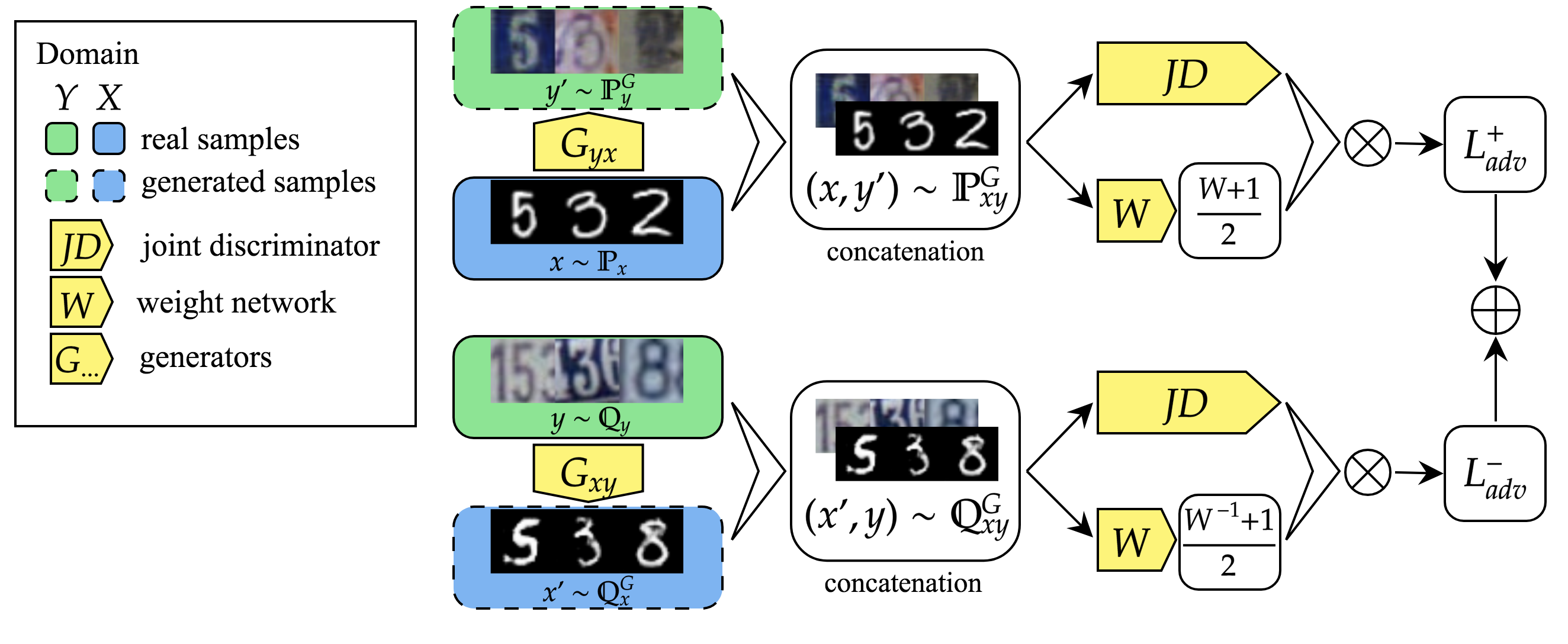}
    \caption{Scheme of the proposed model.} 
    \label{fig:diagram}
\end{figure*}
Alternative solution which might be more robust would involve re-weighting both samples. Thanks to the fact that if Radon-Nikodym derivatives between two distributions exist in both directions they are inverses of each other, we can formulate batch weight so that both distributions are resampled to match at their midpoint $\frac{1}{2}(\PP_y^G + \QQ_y)$. We explore this idea in the following subsection. 
    
    

\subsection{Reweighting both domains}
Here we assume the existence of two joint distributions $\PP_{xy}, \QQ_{xy}$ on $\XX\times\YY$ such that $\textrm{supp }\PP_{xy} = \textrm{supp } \QQ_{xy}$ and that we observe samples drawn from their marginals $\PP_x$ and $\QQ_y$. $\PP_{xy}$ and $\QQ_{xy}$ are said to represent a correct matching between these domain, i.e the transfer from $x$ to $y$ (or other way around) is considered valid if the pair $(x,y)$ can be drawn from $\PP_{xy}$ and $\QQ_{xy}$. Although we do not assume the same marginals, we do assume the equality of conditionals,
\begin{equation}\label{eq:conditionals}
    \PP_{y|x} = \QQ_{y|x}\qquad\textrm{and}\qquad\PP_{x|y} = \QQ_{x|y}.
\end{equation}

We aim to obtain the correct transfer by learning generators $G_{xy}:\XX\to\YY, G_{yx}:\YY\to\XX$ that mimic the above marginals. Let $\PP^G_y = G_{xy}\#\PP_x$ and $\QQ^G_x = G_{yx}\#\QQ_y$.

Let $\mathbb{M} = \frac{1}{2}(\PP_{xy} + \QQ_{xy})$ be a mixture of the two joint distributions. Thanks to the assumption of equality of the supports of $\PP_{xy}$ and $\QQ_{xy}$, both Radon-Nikodym derivatives $w = \frac{d\PP_{xy}}{d\QQ_{xy}}$ and $v = \frac{d\QQ_{xy}}{d\PP_{xy}}$ exist and satisfy,
\begin{equation}\label{eq:inverseRN}
    w(x, y) = \left(v(x, y)\right)^{-1},\qquad (x, y)\in\textrm{supp }\PP_{xy}.
\end{equation}
Therefore
\begin{align} \label{eq:midpoint}
    \PP_{xy}\tfrac{1}{2}(1 + w(X,Y)) &= \mathbb{M} 
    \nonumber\\
    &= \QQ_{xy}\tfrac{1}{2}(1 + w^{-1}(X,Y)).
\end{align}
As we aim to learn the distributions $\PP_{xy}$ and $\QQ_{xy}$ through generators $G_{xy}$ and $G_{yx}$ (implicitly, by learning the conditionals),
\begin{align*}
    \PP_{xy} \approx \PP^G_{xy} :=& (\textrm{id} \otimes G_{yx})\#\PP_x, \nonumber\\
    \QQ_{xy} \approx \QQ^G_{xy} :=& (G_{xy}\otimes \textrm{id})\#\QQ_y. \nonumber
\end{align*}
We can approximate distributions on the left and right hand sides of the Equation \ref{eq:midpoint} using available samples from marginals $\PP_x$ and $\QQ_y$, along with weighting network $W:\XX\times\YY\to\mathbb{R}^+$ that approximates the derivative $w = \frac{d\PP_{xy}}{d\QQ_{xy}}$. 

Therefore, at generation step we should optimize the following objective:
\begin{equation}\label{eq:objective2}
  \inf_{G_{xy}, G_{yx}, W} \EE_{\substack{X\sim\PP_x \\ Y\sim\QQ_y}}\mathcal{L}\left(\PP^G_{xy}\tfrac{1}{2}(1 + W),
  \QQ^G_{xy}\tfrac{1}{2}(1 + W^{-1})\right),
\end{equation}
where $\mathcal{L}$ is some loss function trained adversarially.

At this point, we introduce a \emph{joint discriminator} $D:\XX\times\YY\to R$, a neural network that discriminates between distributions supported on $\XX\times\YY$ and $R$ is a domain dependent on the GAN type. Similar idea has been applied in ALI \citep{ali}. Joint discriminator enforces cycle-consistency on the abstract- rather that pixel-level as objective \ref{eq:cycle-consistency} does.

Assuming yet again Wasserstein setting and $R = \mathbb{R}$, the full objective implied by Eq.  \ref{eq:objective2} is as follows:
\begin{align} \label{eq:objective3}
    &\inf_{G_{xy}, G_{yx}, W} \sup_D \big(
    \nonumber\\
    &\EE_{X\sim\PP_x}\tfrac{1}{2}D(X, G_{yx}(X))
    \times (1 + W(X, G_{yx}(X)))
    \nonumber\\
    -&\EE_{Y\sim\QQ_y}\tfrac{1}{2}D(G_{xy}(Y), Y)
    \times (1 + W(G_{xy}(Y), Y)^{-1})\big).
\end{align}

The batch weight procedure for Wasserstein domain transfer with joint discriminator is detailed in Algorithm~\ref{alg:bw_v2}. Overview of the algorithm is also shown in Figure \ref{fig:diagram}. From now on, we will call the proposed architecture a \emph{Joint Discriminator - Batch Weighted} domain transfer or shortly \emph{JD-BW}.

\begin{algorithm}
   \caption{Batch Weight v2.}
   \label{alg:bw_v2}
\begin{algorithmic}
   \STATE {\bfseries Given:} $\PP_x$ and $\QQ_y$ - source and target distributions 
   \STATE  {\bfseries Given:} $d$ - number of discriminator steps per generator step, $N$ - total training steps, $m$ - batch size
   \STATE Initialize generators $G_{xy}, G_{yx}$, discriminator $D$ and weighting $W$ network parameters $\theta_G, \theta_D, \theta_W$.
   \FOR{$k=1$ {\bfseries to} $n$}
   \STATE \texttt{\# generator - weight step}
   \STATE Sample $x_1,\ldots,x_m\sim\PP$ and $y_1,\ldots,y_m\sim\QQ$.
   \STATE $w_1, \ldots, w_m \leftarrow
   \sigma([W(x_i, G_{yx}(x_i))]_{i=1}^m)$
   \STATE $v_1, \ldots, v_m \leftarrow
   \sigma(-[W(G_{xy}(y_i), y_i)]_{i=1}^m)$
   \STATE $L^- \leftarrow \sum_{i=1}^m D(x_i, G_{xy}(x_i)) \cdot \tfrac{1}{2}(1 + w_i)$
   \STATE $L^+ \leftarrow \sum_{i=1}^m D(G_{yx}(y_i), y_i) \cdot \tfrac{1}{2}(1 + v_i)$
   \STATE $\theta_G \leftarrow \textrm{Adam}\left(\nabla_{G} [L^- - L^+], \theta_G\right)$
   \STATE $\theta_W \leftarrow \textrm{Adam}\left(\nabla_W\left[(L^- - L^+)^2\right], \theta_W\right)$
   \FOR{$j=1$ {\bfseries to} $d$}
    \STATE Sample $x_1,\ldots,x_m\sim\PP$ and $y_1,\ldots,y_m\sim\QQ$.
   \STATE $w_1, \ldots, w_m \leftarrow
   \sigma([W(x_i, G_{yx}(x_i))]_{i=1}^m)$
   \STATE $v_1, \ldots, v_m \leftarrow
   \sigma(-[W(G_{xy}(y_i), y_i)]_{i=1}^m)$
   \STATE $L^- \leftarrow \sum_{i=1}^m D(x_i, G_{xy}(x_i)) \cdot \tfrac{1}{2}(1 + w_i)$
   \STATE $L^+ \leftarrow \sum_{i=1}^m D(G_{yx}(y_i), y_i) \cdot \tfrac{1}{2}(1 + v_i)$
   \STATE $\theta_D \leftarrow \textrm{Adam}\left(-\nabla_D [L^- - L^+], \theta_D\right)$  
   \ENDFOR
   \ENDFOR
\end{algorithmic}
\end{algorithm}

\subsection{Two-domain vs. single-domain batch weight}
The advantage of reweighting both domains is that it ensures that every example in each training set would receive a weight no smaller than a half of what it would get without reweighting. Therefore it may never collapse in the sense that some examples from one domain are wrongly excluded from training, i.e. assigned zero weights.

On the other hand, one-sided reweighting carries a risk of some training examples from the weighted domain getting weights very close to zero, which slows down the training (as they would have small impact on gradients). Some examples may need to be assigned very small weights, yet this can occur wrongly at early phase of training, when weighting network is imperfect. For instance, a possible failure mode would be when such low weights were assigned to the samples of lowest quality (at some point during training); in such case generator would have very little incentive to improve them.

In practice, we found reweighting both domains much more stable. Thanks to the symmetric formulation\footnote{more precisely, the invertibility of Random-Nikodym derivative.}, if some mode gets lower weights in one domain, the corresponding mode in the other domain is likely to receive higher weights, eventually leading to balance between reweighted domains.

\subsection{Non-uniqueness and implicit bias} \label{s:implicit-bias}
It is important to note that given empirical distributions $\PP_x$ and $\QQ_y$ there exist many joints $\PP_{xy}$ and $\QQ_{xy}$ satisfying equality of conditionals (Assumption \ref{eq:conditionals}). For instance, $\PP_{xy}^{0} := \QQ_{xy}^0 := \PP_x\otimes\QQ_y$ is a valid distribution on $\XX\times\YY$ that has all the assumed properties, except that it leads to \emph{independent} transfer between these domains. 

\begin{figure*}
    \centering
    \begin{subfigure}{0.47\linewidth}
        \includegraphics[width=\textwidth]{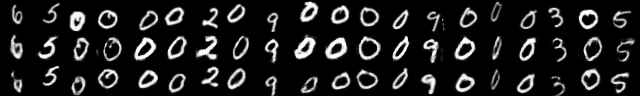}
        \caption{Joint Discriminator with Batch Weight}
        \label{fig:mnist:bw-jd}
    \end{subfigure}
    ~ 
    \begin{subfigure}{0.47\linewidth}
        \includegraphics[width=\textwidth]{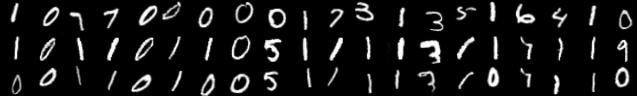}
        \caption{MUNIT with Batch Weight}
        \label{fig:mnist:bw-munit}
    \end{subfigure}
    \\
    \begin{subfigure}{0.47\textwidth}
        \includegraphics[width=\textwidth]{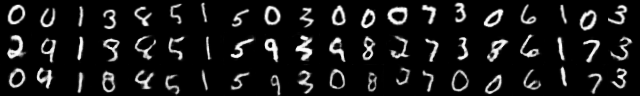}
        \caption{Joint Discriminator alone}
        \label{fig:mnist:jd}
    \end{subfigure}
    ~
    \begin{subfigure}{0.47\textwidth}
        \includegraphics[width=\textwidth]{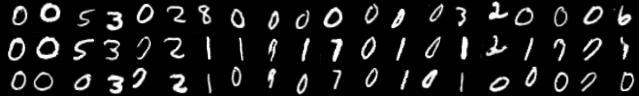}
        \caption{MUNIT}
        \label{fig:mnist:munit}
    \end{subfigure}
    \caption{Results for SR-MNIST to MNIST transfer. Each picture shows three rows of images: original SR-MNIST samples in the first one, samples transferred to MNIST space in the second one, and 2nd row samples transferred back to SR-MNIST space in the last row. Only Joint Discriminator architecture with Batch weight achieves satisfying results, while all other models struggle with frequency of zeros in SR-MNIST dataset, which are often matched with other digits. Note that JD models are do not directly optimize the quality of reconstructions (2nd rows), whereas MUNIT does so via cycle-consistency. }
    \label{fig:mnist}
\end{figure*}
It is worth noting the role of implicit bias of the generator network architectures used in modelling the conditional distributions $\PP_{y|x}$ and $\QQ_{x|y}$. State-of-the art image-to-image translation models \citep{cycle-gan, munit, drit} all use deep ResNets \citep{resnet} or U-Nets \citep{unet} as transfer-generators. These architectures bias the generator mapping towards identity in pixel-level space, which helps obtaining satisfying transfer networks. This, however, also explains limitations of these models: the most impressive performance has so far been achieved in tasks with near pixel-to-pixel correspondence, also referred to as \emph{style transfer}.

Following these approaches, we impose similar architectural constraints on $G_{xy}, G_{yx}$ to enforce the dependence structure in learned joints $\PP^G_{xy}$ and $\QQ^G_{xy}$.

\section{Experiments}
\begin{figure}
    \centering
    \vspace*{-20pt}
    \includegraphics[width=\linewidth]{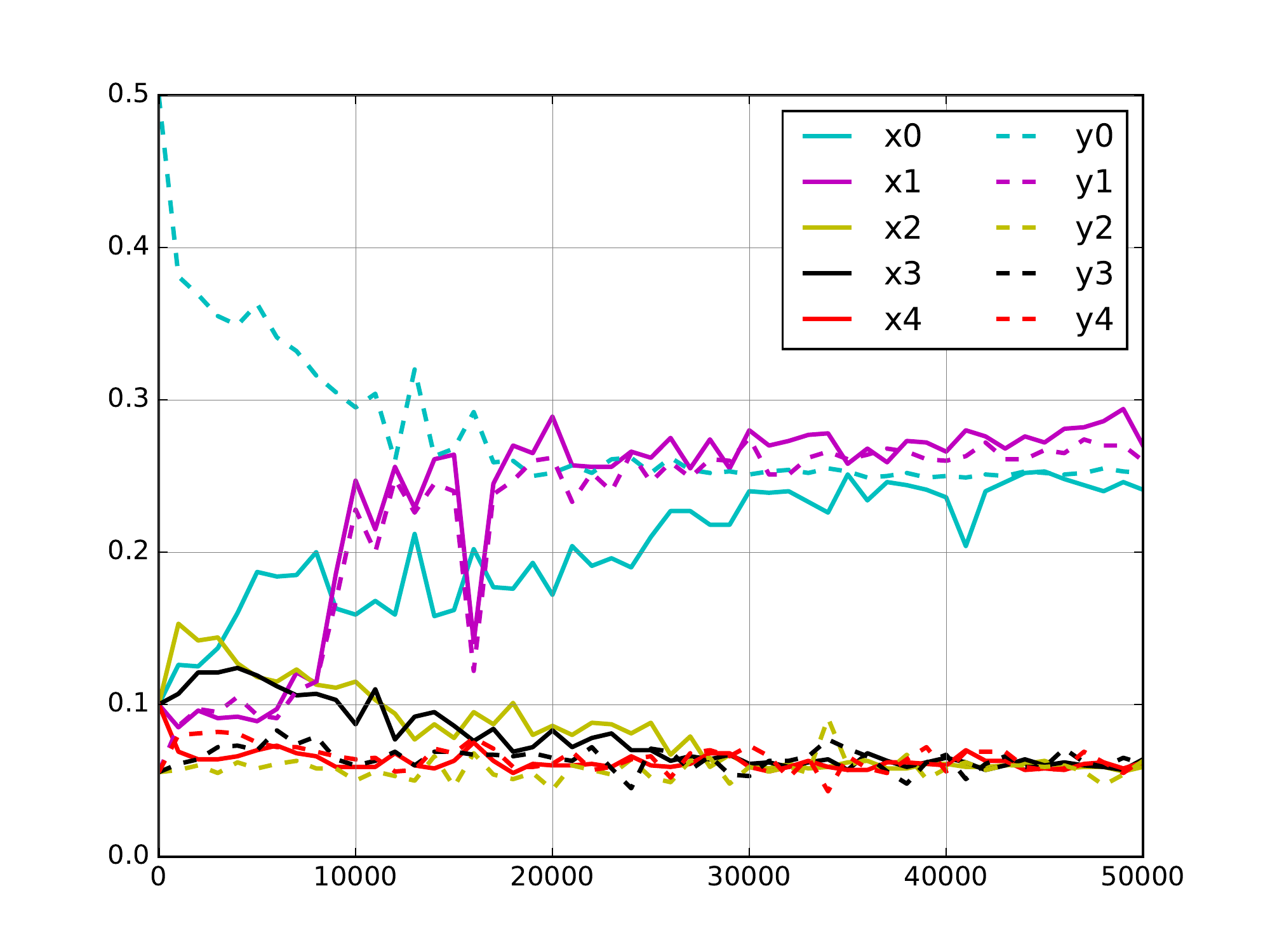}
    \vspace*{-20pt}
    \caption{Moving averages of the combined batch weights assigned to each of the digit-classes throughout MNIST to SR-MNIST transfer training; iterations in horizontal axis. For clarity we show values only for first five digits; $xk$'s - solid lines ($yk$'s - dashed lines) stand for digits $k$ in MNIST (SR-MNIST). Even though zeros in SR-MNIST are very frequent, their weights gradually become lower, while those of their MNIST - counterparts - higher, eventually matching with each other. Total weights of other digits match too, with ones getting high weights in both datasets.}
    \label{fig:mnist:plot}
    \vspace*{-5pt}
\end{figure}
The early empirical studies confirmed the issues mentioned in Section \ref{s:issues}: although re-weighting only generated samples (Algorithm \ref{alg:bw}) often leads to correct transfer it can be unstable. The failure modes that occur in this setting are often caused by the weighting network assigning all weight to a single example in each batch. For these reasons in this Section we focus on experiments with Algorithm \ref{alg:bw_v2}.

\subsection{Datasets}
We carry out experiments on four dataset pairs.
\begin{enumerate}
    \item \textbf{MNIST to skewed \& resized MNIST}. In this experiment we alter the standard MNIST dataset by introducing bias towards zeros. In the \emph{SR-MNIST} (skewed and resized MNIST) half of the samples are drawn from the class of zeros, while the other half are drawn with equal probabilities from the remaining digit classes. The images are then padded, randomly rotated by the angle $\alpha\in(-\frac{\pi}{12},\frac{\pi}{12})$ and randomly cropped. The resulting digits are slightly smaller than the original ones and not necessarily centered. Although changing sampling frequency itself makes the case for batch weighting, the alterations made to the digits so that the transfer to be learnt is not trivial/deterministic.
    
    \item \textbf{MNIST to SVHN}. We tackle the task of transferring between MNIST to SVHN~\citep{svhn} without using the labels. SVHN has an non-uniform distribution of digits and is characterized by considerably more complex features, such us font, colour, background and size. Although we use the version of SVHN with centered digits, they also often contain side-digits coming from the whole house number. 
    To our best knowledge, this problem has not yet been solved. 
    
    \item \textbf{Edges to Shoes\&Bags}. 
    We combine \emph{edges2shoes} and \emph{edges2handbags} datasets \citep{pix2pix} to obtain two-class datasets of edges and photos, and alter sampling of the latter so that 90\% of examples are photos of shoes. In the edge-domain we leave sampling unchanged, hence ~50k out of total ~188k (26\%) examples are contours of shoes. We carry out experiments at 128x128 resolution.
    
    \item \textbf{CelebA to Portraits}. We transfer CelebA dataset of celebrity photos \citet{celeba} to WikiArt dataset of 1714 portraits \citet{drit}. We randomly crop images around the faces and resize to 128x128 resolution. 
\end{enumerate}
\begin{figure}
    \centering
    \includegraphics[width=.65\linewidth]{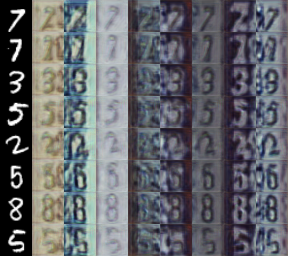}
    
    \caption{MNIST to SVHN transfer with JD-BW architecture and fixed noise values. Original MNIST samples on the left; samples in each other column were obtained with the same noise sampled from $\mathbb{R}^{16}$. }
    \label{fig:mnist-svhn:paired}
    \vspace*{-5pt}
\end{figure}

\subsection{Benchmarks and ablation study}
We compare the performance of the proposed model with MUNIT \citep{munit}, which is one of the state-of-the-art models in unsupervised image-to-image transfer. We do not compare with CycleGAN \citep{cycle-gan} and BiCycleGAN \citep{bicyclegan} as the former does not allow multimodal transfer and MUNIT is essentially its extension, while the latter requires paired training examples. 

Since our model has two novel components, batch weight and joint discriminator, we also carry out two ablations on the MNIST - SR-MNIST task (Section~\ref{s:experiments:mnist-mnist}): 
\begin{itemize}
    \item MUNIT with batch weight,
    \item Joint discriminator architecture without batch weight.
\end{itemize}

\ifarxiv

\else
    To make our research reproducible, the code for experiments is attached to the submission.
\fi

\subsection{Network architectures}

\textbf{Generators}\\
As stressed in Section \ref{s:implicit-bias}, the architecture plays very important role in domain transfer. Following the successful architectures of \citet{cycle-gan, munit} we use generators with several residual blocks \citep{resnet} to bias the transfer towards identity.

Since we consider non-deterministic transfer, generator networks take as inputs the image and noise vector, sampled uniformly from $\mathbb{R}^d$, where $d=8$ for MNIST - SR-MNIST task and $d=16$ for other tasks. Noise vector is repeated over the spacial dimensions and concatenated to convolutional representation: halfway through the depth of the network for 32x32 models and before the first residual block for 128x128 architectures. We follow \citet{biggan} and use spectral normalization \citep{sngan} in both generator and discriminator networks. 32x32 architecture is shown in details in Table \ref{t:generator-architecture} in \ifarxiv Appendix \ref{a:architecture-details}\else supplementary material\fi. For 128x128 resolution we used the same generators as in MUNIT \citep{munit}\footnote{except that we used spectral normalization and did not use \emph{Adaptive Instance Norm}.}.

\textbf{Discriminator}\\
For joint discriminator, we use somewhat more powerful discriminator than the DCGAN \citep{dcgan}, as it has to discriminate between joint distributions on $\XX\times\YY$. The architecture at each level separately computes features of each of the images alone and of their concatenation. We use spectral normalization \citep{sngan} and gradient penalty at training points\footnote{instead of interpolations between training and reference samples as originally proposed by \citet{wgan-gp}} as in \citep{roth-gp}. Details are shown in Table \ref{t:jd-architecture} in \ifarxiv Appendix \ref{a:architecture-details}\else supplementary material\fi.

\textbf{Weighting network}\\
For weighting network, we considered several approaches. Function $W$ maps from $\XX\times\YY$, yet the samples it will ever see during the training are either of the form $(x, G_{yx}(x)), x\sim\PP_x$ or $(G_{xy}(y), y), y\sim\QQ_y$; there are thus multiple ways of modelling $W$. Overall, out of the several strategies we considered for obtaining weights $w_{\xx}, w_{\yy}$ for the batches $\xx\sim\PP_x^n, \yy\sim\QQ_y^n$ the following proved most stable:
\begin{align*}
    W_x&:\XX\to\mathbb{R},\quad W_y:\YY\to\mathbb{R}, \nonumber\\
    w_{\xx} &= \tfrac{1}{2}\left(\sigma(W_x(\xx)) + \sigma(-W_y(G_{xy}(\xx))\right), \nonumber \\
    w_{\yy} &= \tfrac{1}{2}\left(\sigma(-W_x(G_{yx}(\yy)) + \sigma(W_y(\yy))\right), \nonumber
\end{align*}
where the weight networks $W_x, W_y$ are modeled using the architecture of DCGAN discriminator \citep{dcgan} with four convolutional layers and $64$ features in the first layer.

We also found useful regularizing the weighting network training by clipping the values of $W$, which lets us control (and gradually relax) the ratio between highest and lowest weights within a batch. We discuss these further in Appendix \ref{a:weighting}.

\subsection{Training details}
We train all models using Adam optimizer \citep{adam_ref} with parameters $\beta_1=.5, \beta_2=.999$, but we used different hyperparameter settings for 32x32 and 128x128 architectures.

\textbf{32x32 models.} We used batch size of 128. Joint Discriminator models are trained with 5 discriminator steps per generator step, while MUNIT (benchmark) models are trained with one generator per one discriminator step, as in original implementation. For these reason, we train the latter for 3x more generator steps than the proposed architectures\footnote{This required slightly longer training for MUNIT anyway, due to simplified JD architecture.}. After this number of iterations, MUNIT models seemed to converge. Overall, we train JD (MUNIT) for 50k (150k) steps in MNIST - SR-MNIST task and 250k (750k) steps in MNIST - SVHN task.

\textbf{128x128 models}. We used batch size of 6, 2 discriminator steps per one generator step and trained for 300k generator steps.

\section{Results}
\begin{figure*}
    \centering
    \begin{subfigure}{0.47\linewidth}
        \includegraphics[width=\textwidth]{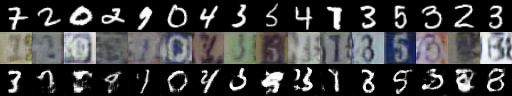}
        \caption{JD-BW, MNIST to SVHN}
        \label{fig:jd-mnist-svhn}
    \end{subfigure}
    ~ 
    \begin{subfigure}{0.47\linewidth}
        \includegraphics[width=\textwidth]{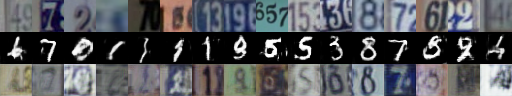}
        \caption{JD-BW, SVHN to MNIST}
        \label{fig:jd-svhn-mnist}
    \end{subfigure}
    \\
    \begin{subfigure}{0.47\textwidth}
        \includegraphics[width=\textwidth]{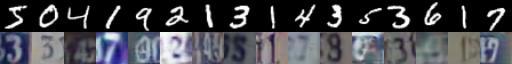}
        \caption{MUNIT, MNIST to SVHN}
        \label{fig:munit-mnist-svhn}
    \end{subfigure}
    ~
    \begin{subfigure}{0.47\textwidth}
        \includegraphics[width=\textwidth]{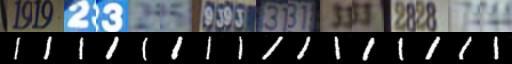}
        \caption{MUNIT, SVHN to MNIST}
        \label{fig:munit-svhn-mnist}
    \end{subfigure}
    \caption{Results for MNIST to SVHN transfer. First row in each picture shows original samples, while second the transferred samples. For the proposed JD-BW model, the additional 3rd row presents samples transferred back to the original space. Although MUNIT produces sharper images, the match between original and transferred samples is very poor, with SVHN to MNIST direction suffering from mode collapse. Joint Discriminator struggles with image sharpness, however the matching is usually correct and cycle-consistent.}\label{fig:mnist-svhn}
\end{figure*}
\begin{figure*}
    \centering
    \begin{subfigure}{0.47\linewidth}
        \includegraphics[width=\textwidth]{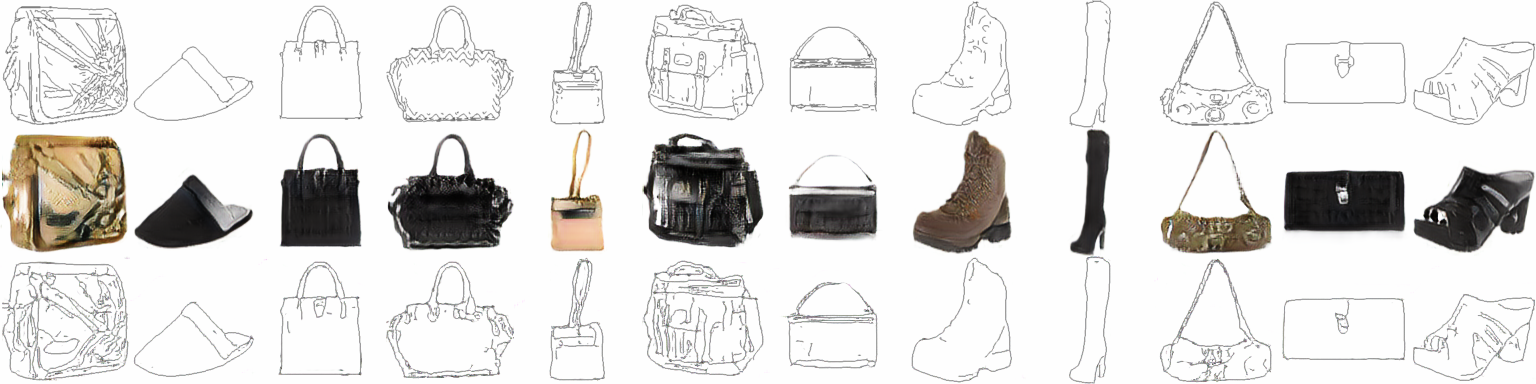}
        \caption{JD-BW (ours), Edges to Shoes\&Bags}
        \label{fig:edges:bw-jd}
    \end{subfigure}
    ~ 
    \begin{subfigure}{0.47\linewidth}
        \includegraphics[width=\textwidth]{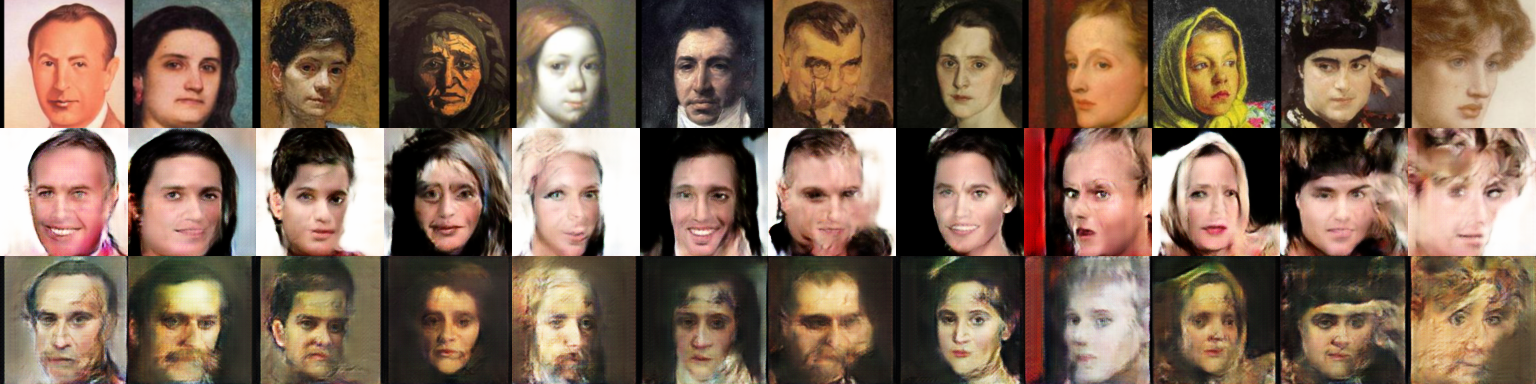}
        \caption{JD-BW (ours), Portraits to CelebA}
        \label{fig:portrait:bw-jd}
    \end{subfigure}
    \\
    \begin{subfigure}{0.47\textwidth}
        \includegraphics[width=\textwidth]{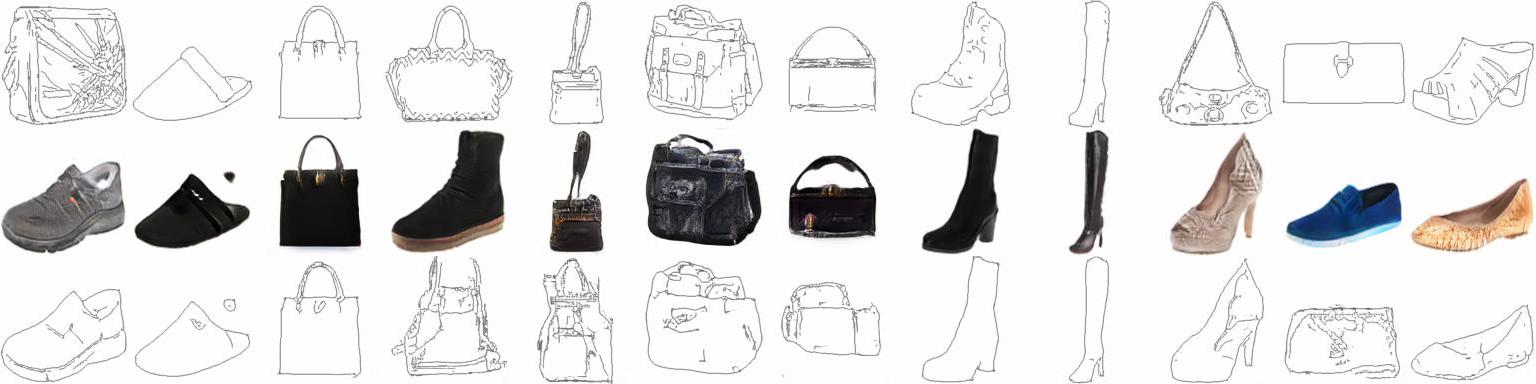}
        \caption{MUNIT, Edges to Shoes\&Bags}
        \label{fig:edges:munit}
    \end{subfigure}
    ~
    \begin{subfigure}{0.47\textwidth}
        \includegraphics[width=\textwidth]{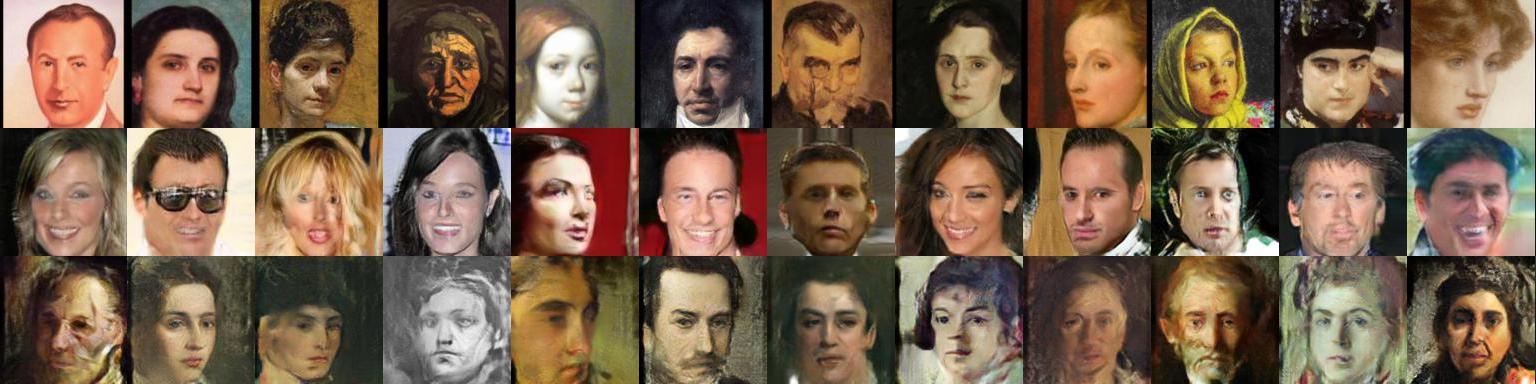}
        \caption{MUNIT, Portraits to CelebA}
        \label{fig:portrait:munit}
    \end{subfigure}
    \caption{Results for Portraits to CelebA and Edges to Shoes\&Bags transfers. Original samples are shown in the first row, second one shows samples transferred to the other domain while the third shows 2nd row transferred back to the original domain. For both datasets our model yields very reasonable transfer. MUNIT, on the other hand, struggles to match domains correctly, despite overall good quality of produced samples. }\label{fig:portrait_and_edges}
\end{figure*}
\subsection{MNIST to SRMNIST} \label{s:experiments:mnist-mnist}
We perform the unsupervised transfer task from SR-MNIST to MNIST  (i.e., without using the labels in any part of our objective for any network). As zeros are over-represented in the first dataset, we anticipate this task will be difficult without properly reweighting the GAN objective. 
We compare to Multimodal Unsupervised Image-to-Image Translation~\citep[MUNIT,][]{munit}

Results from this experiment are shown in Figure~\ref{fig:mnist}. Only Joint Discriminator architecture with Batch Weight performed well, correctly matching different digit classes. Other models often incorrectly match some kinds of SR-MNIST zeros with other MNIST digits. We monitored the batch weights assigned to each example within a batch in order to find out if weighting network(s) are capable of matching frequencies of the modes in two distributions. Figure \ref{fig:mnist:plot} presents evolution of the weights aggregated for each of the MNIST/SR-MNIST classes (as these are the only modes we can clearly distinguish). Weighting network successfully allows the mode frequencies to gradually match between both distributions.

We also note that the proposed model achieves cycle-consistency in an abstract, high-level sense. The learned transfer is non-deterministic, yet the reconstructed SR-MNIST samples $G_{xy}(G_{yx}(y))$ belong to the same sub-manifold as the original samples $y$, where such sub-manifold is spanned by features non-existent in MNIST space.

\subsection{MNIST to SVHN}
Results from this experiment are shown in the Figure~\ref{fig:mnist-svhn}. Although our model does not produce as sharp images as expected and makes few mismatch errors, it provides reasonable transfer. MUNIT, on the other hand, wrongly transfers MNIST to SVHN and collapses completely in the opposite direction.

Our method also disentangles SVHN-specific features from those shared with MNIST, and models the former via noise input. Figure \ref{fig:mnist-svhn:paired} shows samples obtained from different MNIST digits with the same noise, for 8 different noise values.

\subsection{Edges to Shoes\&Bags}
In this experiment the main difficulty comes from difference in frequencies of bags and shoes between source and target domains. As shown in Fig. \ref{fig:edges:bw-jd}, the proposed model successfully tackled the problem, producing correctly transferred samples. MUNIT (Fig. \ref{fig:edges:munit}), on the other hand, struggled with the imbalance and often transferred bag-edges to shoes, which were over-represented in the photo domain.

\subsection{CelebA to Portraits}
We present results from this experiment in Fig. \ref{fig:portrait:bw-jd} and \ref{fig:portrait:munit}. In this experiment we were not able to quantify mass shift between different modes, as no labels/additional features are available for the Portrait dataset. However, some of such imbalances are visible, e.g. gender proportions and frequency of moustache are different in paintings than in celebrity photos.

Both models produced samples of good quality, however those coming from our model preserved much more features of the original examples. MUNIT, in fact, preserves only a pose, while all other features seem to be independent of the source image. In Appendix \ref{a:failure-munit} we further discuss the nature of transfer learned by MUNIT in this and in the previous task.

\section{Conclusion}
In this work, we considered unsupervised domain transfer in the presence of mode imbalance, a situation when modes to be matched have different frequencies in source and target domains. The contributions of this paper are threefold. Firstly, we provide probabilistic formalism of unsupervised domain transfer. Secondly, we propose a novel method of batch weighting to tackle the issue of mode imbalance. Lastly, we propose a new architecture called Joint Discriminator, that not only largely simplifies the training objective, but also ensures cycle-consistency in multi-modal, high-level sense, without directly enforcing quality of reconstructions. We experimentally show effectiveness of our model and its superiority over existing benchmark in tasks where mode-mass imbalance is present.

\ifarxiv
    \subsubsection*{Acknowledgements}
    Authors thank NVIDIA for donating a DGX-1 computer used in this work. M.B. thanks Engineering and Physical Sciences Research Council (EPSRC) for
    funding part of this research.
\else
\fi

\bibliographystyle{IEEEtranN}
\bibliography{binek.bib}

\newpage
\clearpage 
\begin{appendices}

\section{One-sided batch weight - details} \label{a:one-sided}
In Algorithm \ref{alg:bw} we present the original idea of one-sided batch-weighted domain transfer.
\begin{algorithm}
   \caption{Batch Weight}
   \label{alg:bw}
\begin{algorithmic}
   \STATE {\bfseries Given:} $\PP_x$ and $\QQ_y$ - source and target distributions 
   \STATE  {\bfseries Given:} $d$ - number of discriminator steps per generator step, $N$ - total training steps, $m$ - batch size
   \STATE Initialize generator $G$, discriminator $D$ and weighting $W$ networks' parameters $\theta_G, \theta_D, \theta_W$.
   \FOR{$k=1$ {\bfseries to} $n$}
   \STATE \texttt{\# generator - weight step}
   \STATE Sample $x_1,\ldots,x_m\sim\PP_x$ and $y_1,\ldots,y_m\sim\QQ_y$.
   \STATE $w_1, \ldots, w_m \leftarrow
   softmax(W(x_1),\ldots, W(x_m))$
   \STATE $L^- \leftarrow \sum_{i=1}^m D(G(x_i)) \cdot w_i$
   \STATE $L^+ \leftarrow \sum_{i=1}^m D(y_i) \cdot \frac{1}{m}$
   \STATE $\theta_G \leftarrow \textrm{Adam}\left(\nabla_{G} L^-, \theta_G\right)$
   \STATE $\theta_W \leftarrow \textrm{Adam}\left(\nabla_W\left[(L^- - L^+)^2\right], \theta_W\right)$
   \FOR{$j=1$ {\bfseries to} $d$}
    \STATE Sample $x_1,\ldots,x_m\sim\PP_x$ and $y_1,\ldots,y_m\sim\QQ_y$.
   \STATE $w_1, \ldots, w_m \leftarrow
   softmax(W(x_1),\ldots, W(x_m))$
   \STATE $L \leftarrow \sum_{i=1}^m D(G(x_i)) \cdot w_i - \sum_{i=1}^{m} D(y_i)\cdot\frac{1}{m}$
   \STATE $\theta_D \leftarrow \textrm{Adam}\left(-\nabla_D L, \theta_D\right)$  
   \ENDFOR
   \ENDFOR
\end{algorithmic}
\end{algorithm}

\section{Architecture details} \label{a:architecture-details}
Tables \ref{t:generator-architecture} and \ref{t:jd-architecture} present generator and joint-discriminator architectures used in experiments with 32x32 images.
\begin{table}
    \centering
    \begin{tabular}{ccc}
         image $x$ & $\textrm{concat}(x,y)$ & image $y$ \\
         \cmidrule(lr){1-1}\cmidrule(lr){2-2}\cmidrule(lr){3-3}
         4x4 conv(32)  &  4x4 conv(64)&  4x4 conv(32) \\
         $(x_1)$ & ($xy_1$) & ($y1$) \\
         & concat$(x_1, xy_1, y_1)$ & \\
         \cmidrule(lr){1-1}\cmidrule(lr){2-2}\cmidrule(lr){3-3}
         4x4 conv(64)&  4x4 conv(128)&  4x4 conv(64) \\
         ($x_2$)& ($xy_2$) & ($xy_2$) \\
         & concat$(x_2, xy_2, y_2)$ & \\
         \cmidrule(lr){2-2}
         & 2 x ResBlock(128) & \\
         \cmidrule(lr){1-1}\cmidrule(lr){2-2}\cmidrule(lr){3-3}
         4x4 conv(128)&  4x4 conv(256)&  4x4 conv(128) \\
         ($x_3$)& ($xy_3$)& ($y_3$) \\
         & concat$(x_3, xy_3, y_3)$ & \\ 
         \cmidrule(lr){1-3}
         & 4x4 conv(256) & \\
         \cmidrule(lr){2-2}
         & fc 1024 $\to$ 256 & \\
         \cmidrule(lr){2-2}
         & fc 256 $\to$ 1 & \\         
    \end{tabular}
    \caption{Joint discriminator architecture. Each convolution has stride 2. Residual blocks \citep{resnet} contain two 3x3 convolutions and skip connection.}
    \label{t:jd-architecture}
\end{table}
\begin{table}
    \centering
    \begin{tabular}{cc}
         image $x\in\mathbb{R}^{c\times32\times32}$ & noise $z\in\mathbb{R}^d$ \\
         \cmidrule(lr){1-1}\cmidrule(lr){2-2}
         KxK conv(64), stride $s$  & repeat $(32/s \times 32/s)$ \\
         \cmidrule(lr){1-1}
         2 x ResBlock(64) & ($z'$)\\
         \cmidrule(lr){1-1}
         1x1 conv(64), stride $1$ \\
         ($x'$) & \\
         \cmidrule(lr){1-2}
         \multicolumn{2}{c}{concat$(x', z')$} \\
         \multicolumn{2}{c}{1x1 conv(64), stride $1$} \\
         \cmidrule(lr){1-2}
         \multicolumn{2}{c}{2 x ResBlock(64)} \\
         \cmidrule(lr){1-2}
         \multicolumn{2}{c}{KxK transposed conv(c), stride s} \\
    \end{tabular}
    \caption{Generator network architecture. $c$ denotes number of channels (1 for greyscale, 3 for rgb), $s$ stride and $K$ kernel size. For MNIST - SR-MNIST task $s=2, K=4$, for MNIST - SVHN $s=1, K=1$. Residual blocks \citep{resnet} contain two 3x3 convolutions and skip connection.}
    \label{t:generator-architecture}
\end{table}

\subsection{Weighting network} \label{a:weighting}
We considered three different ways of modelling weight network $W$. Given batches of pairs of real and generated samples $(\xx, G_{yx}(\xx)), \xx\sim\PP_x^n$ and $(G_{xy}(\yy), \yy), \yy\sim\QQ_y^n$ we may get the weights $w_{\xx}, w_{\yy}$ using each of the following architectures
\begin{enumerate}
    \item ($W$ concatenates two arguments)
    \begin{align*}
        W&:\XX\times\YY\to\mathbb{R} \nonumber\\
        w_{\xx} &= \sigma(W(\xx,G_{xy}(\xx))),\quad w_{\yy} = \sigma(-W(G_{yx}(\yy), \yy)). \nonumber
    \end{align*}
    \item ($W$ takes one argument)
    \begin{align*}
        W&:\XX\to\mathbb{R}, \nonumber\\
        w_{\xx} &= \sigma(W(\xx)),\quad
        w_{\yy} = \sigma(-W(G_{yx}(\yy))). \nonumber
    \end{align*}
    \item (composite)
    \begin{align*}
        W_x&:\XX\to\mathbb{R},\quad W_y:\YY\to\mathbb{R}, \nonumber\\
        w_{\xx} &= \tfrac{1}{2}\left(\sigma(W_x(\xx)) + \sigma(-W_y(G_{xy}(\xx)))\right), \nonumber \\
        w_{\yy} &= \tfrac{1}{2}\left(\sigma(-W_x(G_{yx}(\yy))) + \sigma(W_y(\yy))\right). \nonumber
    \end{align*}
\end{enumerate}
The weight network(s) $W$ ($W_x, W_y$) were the same as DCGAN discriminator \citep{dcgan} with four convolutional layers and $64$ features in the first layer.

We found the last (composite) architecture to be the most stable one. The first approach, although the most natural, most probably suffers from the fact that it takes longer for joint samples to look similar to each other than it does for the marginals.

\section{Role of the noise term} \label{a:failure-munit}
\begin{figure*}
    \centering
    \begin{subfigure}{0.47\linewidth}
        \includegraphics[width=\textwidth]{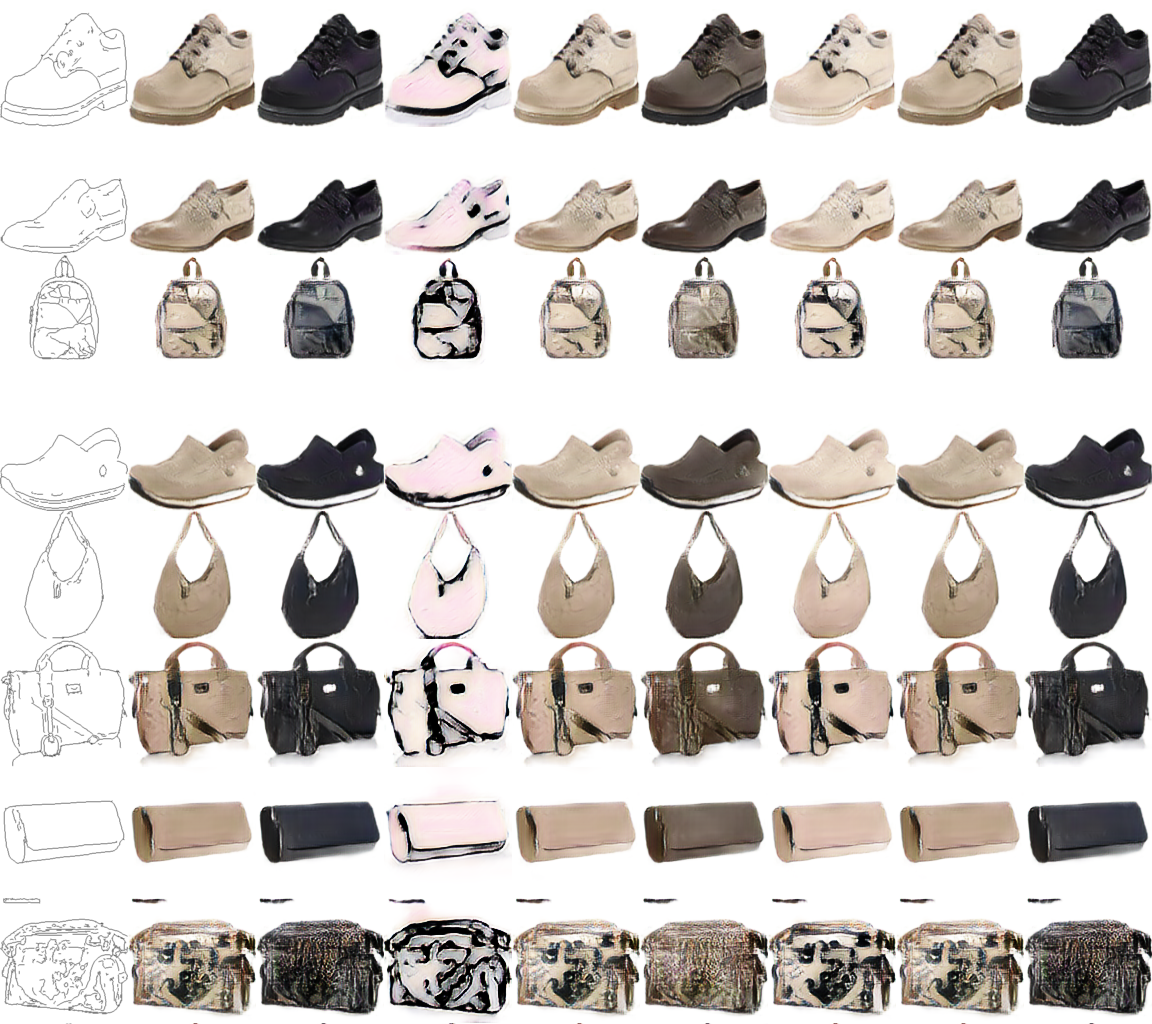}
        \caption{JD - BW (ours)}
        \label{fig:edge-paired:jd-bw}
    \end{subfigure}
    ~ 
    \begin{subfigure}{0.47\linewidth}
        \includegraphics[width=\textwidth]{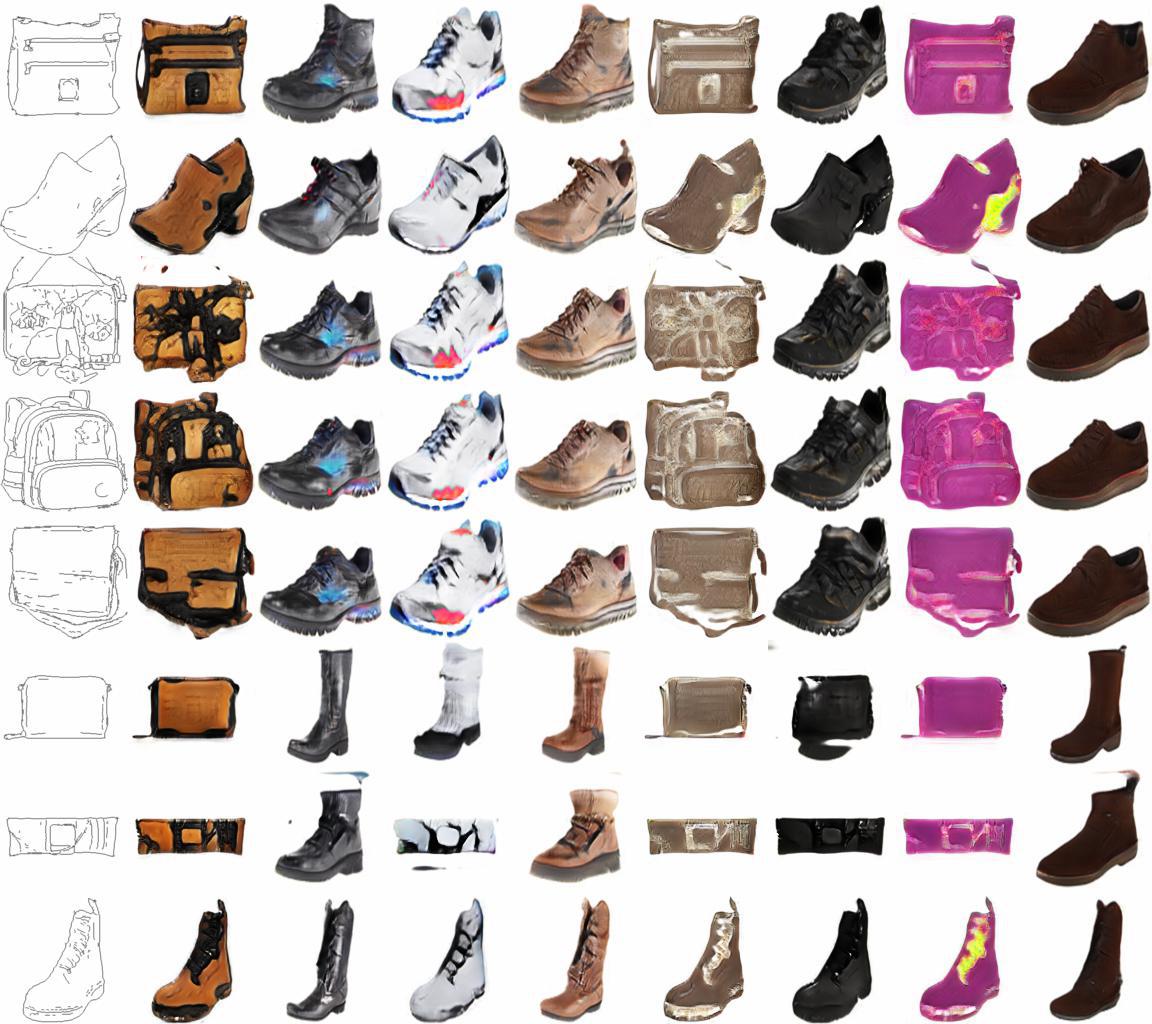}
        \caption{MUNIT}
        \label{fig:edge-paired:munit}
    \end{subfigure}
    \\
    \begin{subfigure}{0.47\textwidth}
        \includegraphics[width=\textwidth]{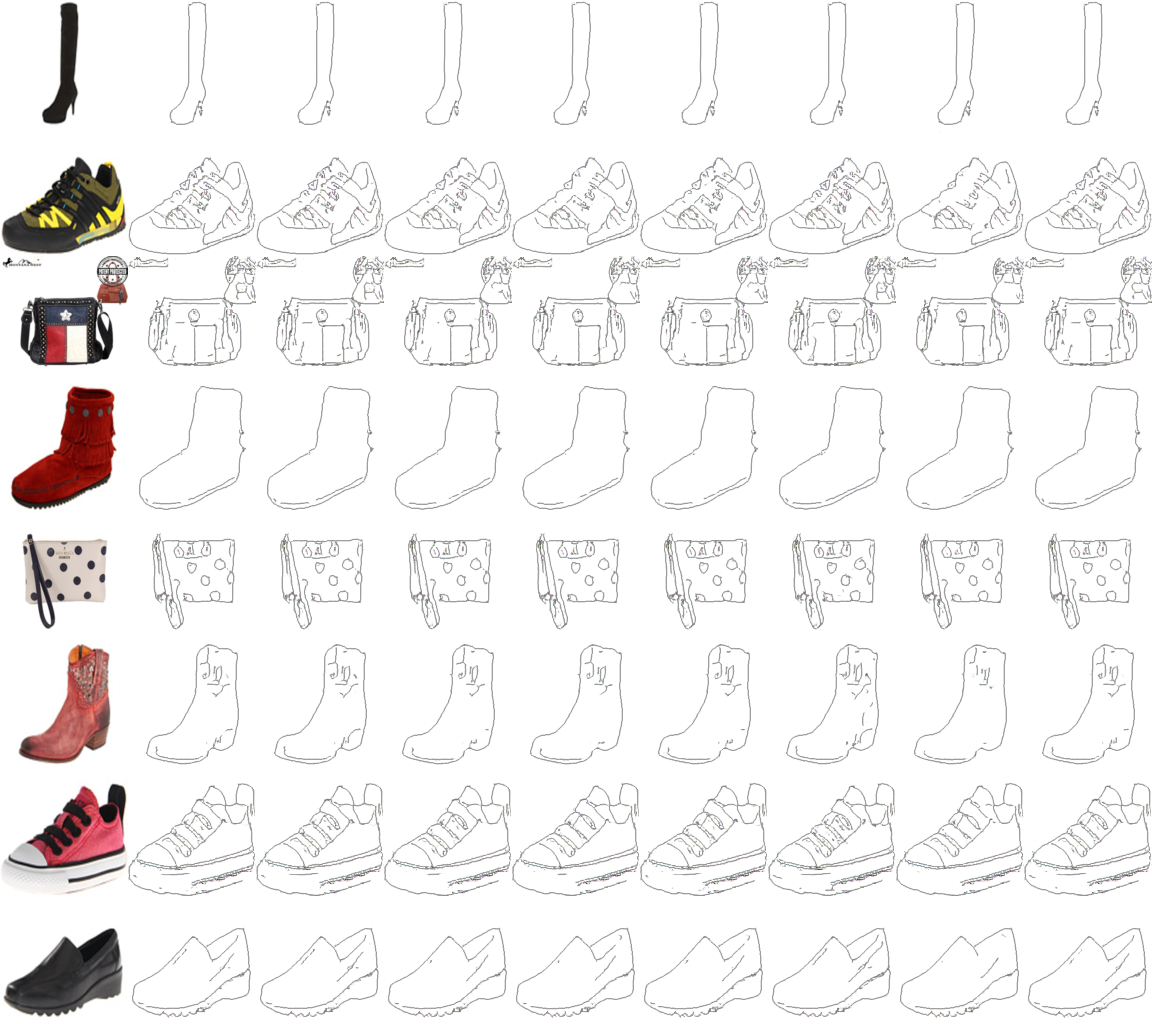}
        \caption{JD - BW (ours)}
        \label{fig:photo-paired:jd-bw}
    \end{subfigure}
    ~
    \begin{subfigure}{0.47\textwidth}
        \includegraphics[width=\textwidth]{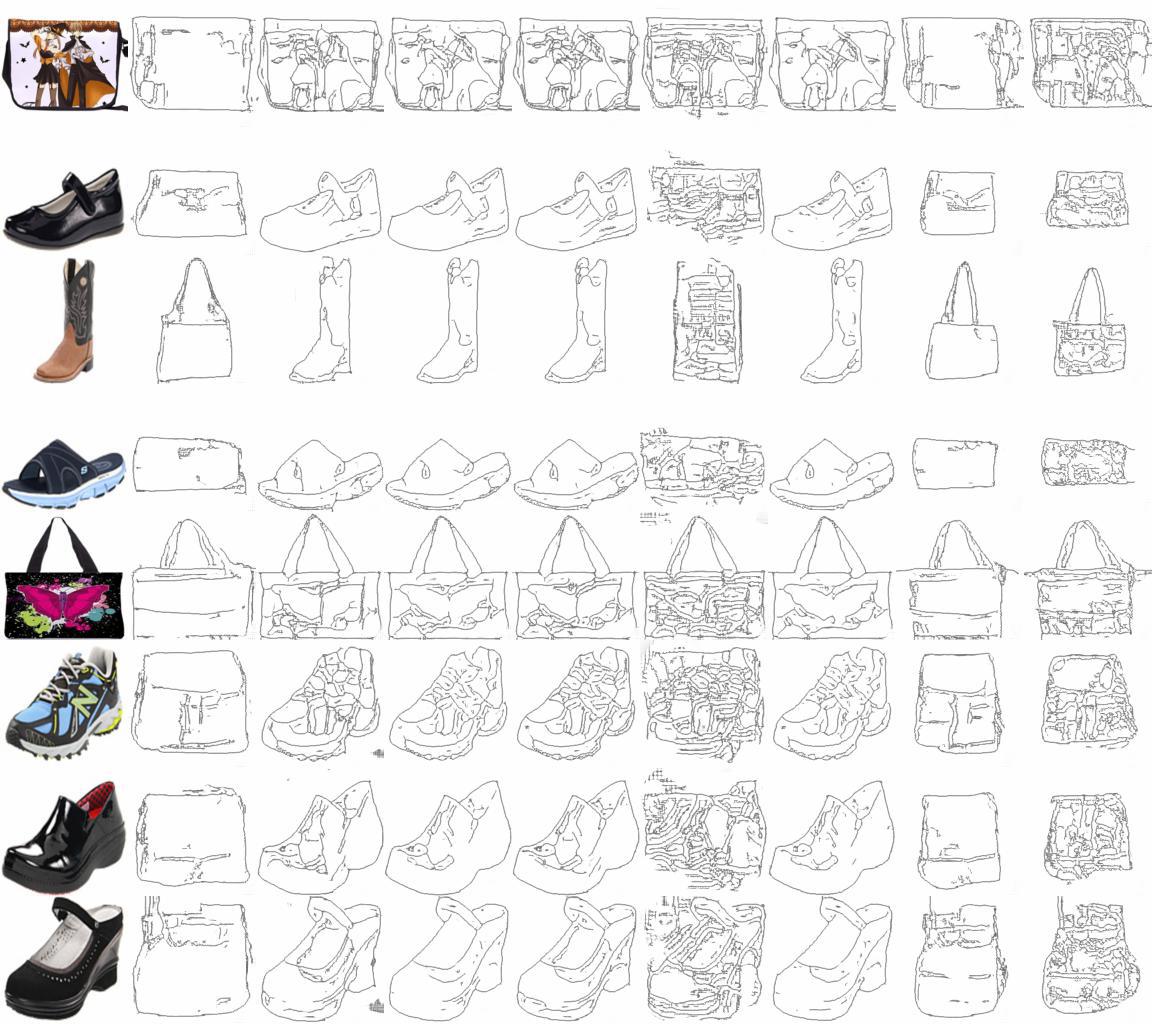}
        \caption{MUNIT}
        \label{fig:photo-paired:munit}
    \end{subfigure}
    \caption{Edges to Shoes{\&}Bags transfer with fixed noise values. In each picture first column represents original images; other columns present transfer with noise term fixed per column and source picture fixed per row. In MUNIT, some noise terms (e.g. columns 3,4,5 in (b); 2, 6, 8 and 9 in (d)) 'deactivate' source images to produce images from over-represented class in the target domain. }\label{fig:edge2photo-paired}
\end{figure*}
\begin{figure*}
    \centering
    \begin{subfigure}{0.47\linewidth}
        \centering
        \includegraphics[width=.8\textwidth]{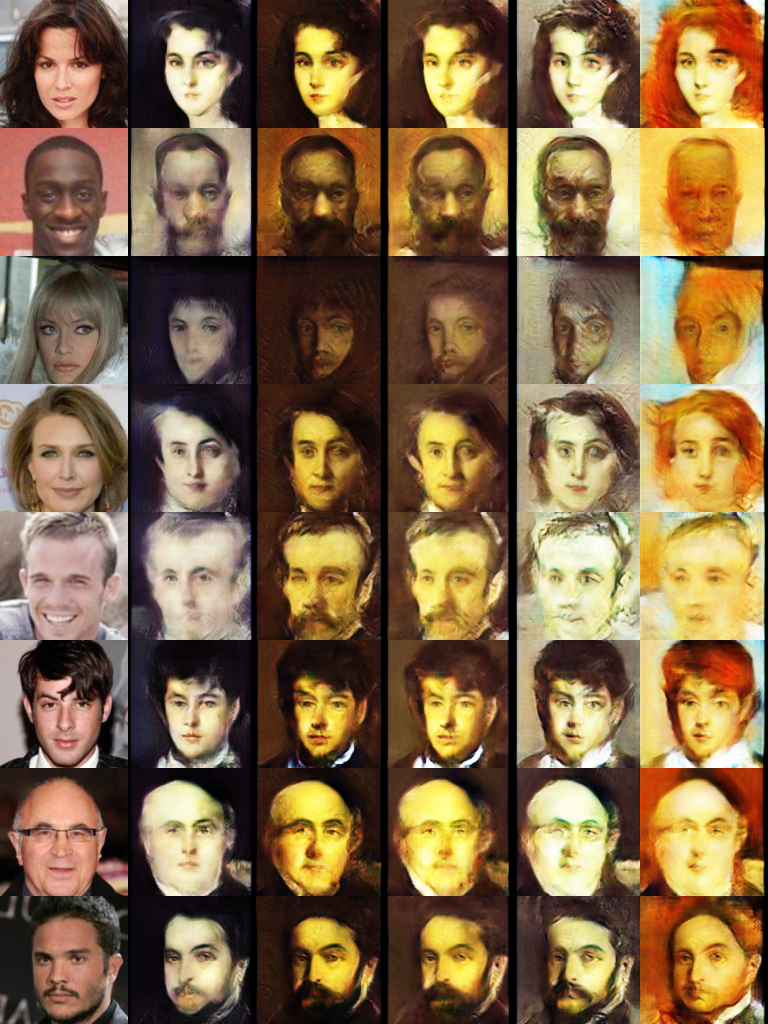}
        \caption{JD - BW (ours)}
        \label{fig:celeba2portrait-paired:jd-bw}
    \end{subfigure}
    ~ 
    \begin{subfigure}{0.47\linewidth}
        \centering
        \includegraphics[width=.8\textwidth]{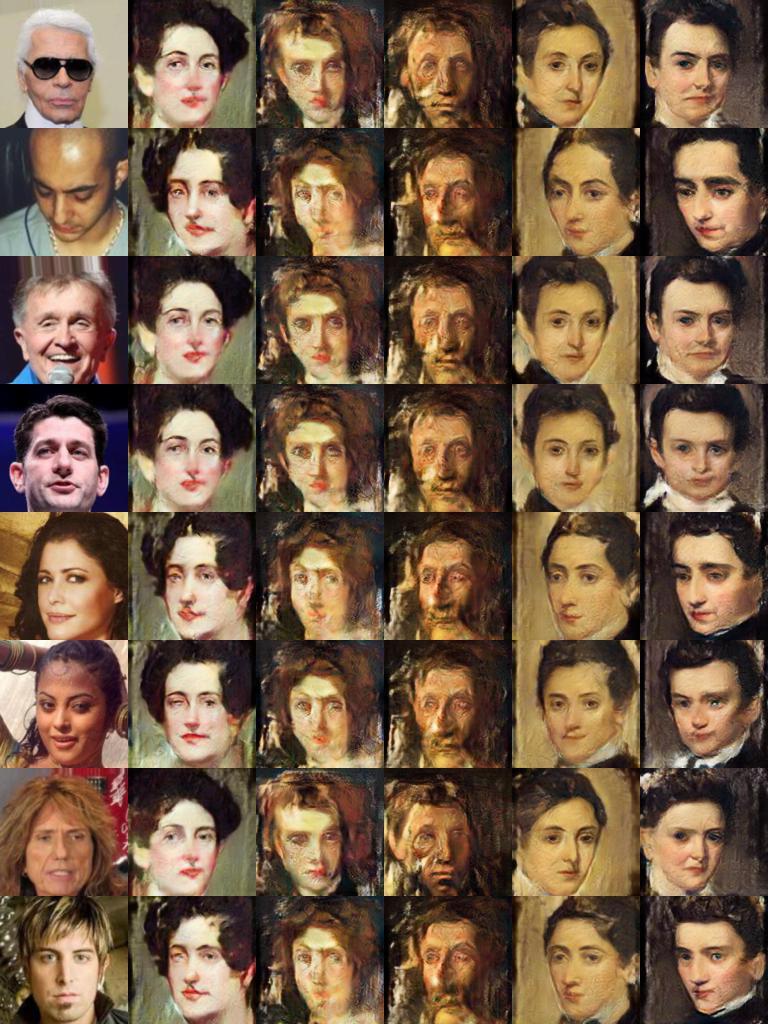}
        \caption{MUNIT}
        \label{fig:celeba2portrait-paired:munit}
    \end{subfigure}
    \caption{CelebA to Portrait transfer with fixed noise values. In both pictures the first columns represent original images; other columns present transfer with noise term fixed per column and source CelebA photo fixed per row. The proposed JD-BW model retains much more facial features, using noise term to encode mostly the portrait style. MUNIT, on the other hand, infers most of the features from noise term, keeping only the position from the source photo. }\label{fig:celeba2portrait-paired}
\end{figure*}
We have observed two types of failures made by MUNIT-trained models in the presence of mode-mass imbalance, both related to what these models encode in the noise term.

In the first one, the class/mode is kept in a transferred sample depending on the noise term. This has been observed in Edges to Shoes\&Bags task, see Figure \ref{fig:edge-paired:munit} and \ref{fig:photo-paired:munit}. In multimodal domain transfer, noise term should only encode the features which are not present in the source domain. In this task, however, the MUNIT-trained model encoded the conditional mode: some noise values caused the generator 'forget' the source image and generate one over-represented in the target domain (regardless of the source image mode/class).

The second issue is the amount of the source image features retained in the transferred one. In CelebA to Portrait transfer, MUNIT tends to keep very few features of the source image (e.g. position of eyes and and nose) and model all high-level features using noise term. As shown in Figure \ref{fig:celeba2portrait-paired:munit}, samples obtained with the same source image are less similar to each other than those generated using the same noise term. With JD-BW model (Fig. \ref{fig:celeba2portrait-paired:jd-bw}) it is the opposite: generators retain much more features of the source image, while the noise terms determines only the style of the generated portraits.

\end{appendices}
\end{document}